\documentclass[runningheads]{llncs}
\usepackage[T1]{fontenc}
\usepackage{graphicx}
\usepackage{amssymb,amsmath,array}
\usepackage{latexsym}
\usepackage{times}
\usepackage{booktabs}
\usepackage{paralist}
\usepackage{subcaption}
\usepackage{hyperref}

\begin{document}
\title{Intelligent Learning Rate Distribution to reduce Catastrophic Forgetting in Transformers}
\titlerunning{Intelligent Learning Rate for Transformers}
%
\author{Philip Kenneweg \inst{1}\orcidID{0000-0002-7097-173X}  \and
Alexander Schulz\inst{1}\orcidID{0000-0002-0739-612X} \and
Sarah Schröder\inst{1}\orcidID{0000-0002-7954-3133}  \and
Barbara Hammer \inst{1}\orcidID{0000-0002-0935-5591}}
\authorrunning{Kenneweg et al.}
%
\institute{¹ Bielefeld University, Inspiration 1, 33615 Bielefeld - Germany}
\maketitle              

\begin{abstract}
Pretraining language models on large text corpora is a common practice in natural language processing. Fine-tuning of these models is then performed to achieve the best results on a variety of tasks.
In this paper, we investigate the problem of catastrophic forgetting in transformer neural networks and question the common practice of fine-tuning with a flat learning rate for the entire network in this context. We perform a hyperparameter optimization process to find learning rate distributions that are better than a flat learning rate. We combine the learning rate distributions thus found and show that they generalize to better performance with respect to the problem of catastrophic forgetting. We validate these learning rate distributions with a variety of NLP benchmarks from the GLUE dataset. 

\end{abstract}

\section{Introduction}
\label{sec:intro}
%

In the real world, data is sequential, tasks may switch unexpectedly and individual tasks may not reoccur for a long time. Thus, the ability to learn tasks in succession is a core component of biological and
artificial intelligence \cite{DBLP:journals/corr/KirkpatrickPRVD16}. Catastrophic forgetting constitutes one major problem in this context \cite{DBLP:journals/corr/abs-0712-3329}: While learning, there is a high risk of forgetting previously learnt information. There has been a large corpus of work dedicated to dealing with this problem \cite{DBLP:journals/corr/abs-1909-08383}. However the problem of catastrophic forgetting also occurs during fine-tuning in modern transformer models \cite{ramasesh2022effect}, yet it is widely unsolved how to efficiently deal with it, with minimal effort. In this approach we propose a comparatively efficient, yet powerful novel approach how to deal with this issue: Careful and automated tuning of the learning rates of a specific network per layer. It turns out that a highly specific and non-monotonic choice obtained via automated bayesian optimization is the best choice, as we will show for state of the art transformer models.

The transformer architecture pioneered by Vaswani et al. \cite{DBLP:journals/corr/VaswaniSPUJGKP17} has enabled large pre-trained neural networks to efficiently tackle previously difficult Natural Language Processing (NLP) tasks with relatively few training examples.
Most language models are pre-trained on a large corpus of text data (for example Wikipedia, Reddit, etc.) using a variety of different unsupervised pre-training objectives (Masked Language Modeling, Next Sentence Prediction, etc.) \cite{bert}. 
In addition, many common architectures use a fine-tuning step for a specific task in parallel with a shallow layer on top of the generated contextualized embeddings to achieve good performance \cite{DBLP:journals/corr/abs-1911-03437}.

In this paper we take a closer look at different kinds of data shifts and how they affect the performances of transformer based networks. Typically, neural networks perform poorly when the data they are trained on is input in a sequential fashion. 
To mitigate this problem, we investigate different learning rate distributions over the layers of the network to decrease the effect of this phenomenon called catastrophic forgetting specifically in transformer based networks.

We hypothesize that different layers of the transformer network represent different abstract concepts and, therefore, should be adapted with different speed when fine-tuning to a new task to reduce catastrophic forgetting. Thereby, we hope to increase the generalization capabilities. We investigate this concept for a few datasets from the NLP domain and compare it to the state of the art from the literature \cite{DBLP:journals/corr/KirkpatrickPRVD16}.

Based on current best practices, manual effort should be minimized; therefore, we aim for an automation of these optimization steps w.r.t\ learning rate distribution. We achieve this by three contributions: first, we define a landscape of promising options for learning rate choices. Second, we introduce an automatic hyperparameter optimization process often referred to as a type of AutoML \cite{HE2021106622}, which enables us to automatically search these options to find the best possible learning rate distribution. Third, we find a good way to combine the results of the optimization into a universally applicable solution.

The source code is open-source and free software, available at \newline \url{https://github.com/TheMody/NAS-CatastrophicForgetting}.


\label{sec:related_work}

\section{Related Work}

At present, the most common approach to fine-tuning a language model is to process the outputs of the transformer with a single layer neural network  \cite{bert,DBLP:journals/corr/VaswaniSPUJGKP17} and fine-tune this stack with a flat learning rate. One of the challenges which occur while fine-tuning is the well known problem of catastrophic forgetting, where the network over-fits on new data and/or forgets previously learned information. This is due to its inherent complexity and comparatively small size of the fine-tuning datasets.
Some recent papers have attributed large pre-trained networks to be affected less by catastrophic forgetting \cite{ramasesh2022effect}. However, they are still affected especially when they are smaller or not as extensively pre-trained, as we will exemplary see later.

In the last years, more sophisticated fine-tuning approaches have evolved to improve this baseline approach.
One wide spread technique to prevent catastrophic forgetting is elastic weight consolidation (EWC) \cite{DBLP:journals/corr/KirkpatrickPRVD16}. Here, an additional loss term is added when fine-tuning the network. This additional loss term prevents the loss of previously learned information by taking the pre-training task into account. During the evaluation section we compare our method to EWC, as it is easy to implement and widely used.

The approach SMART \cite{DBLP:journals/corr/abs-1911-03437} addresses the problem of catastrophic forgetting by introducing a smoothness inducing regularization technique and an optimization method, which prevents aggressive updating of the network weights. While SMART claims to increase performance during fine-tuning with their method, no experiments regarding catastrophic forgetting with consecutive task learning are mentioned in their paper and no code to make evaluations on other settings possible has been published.

Other works have tried to find an optimal learning rate distribution but only for older language models and their respective architectures namely LSTMs. \cite{DBLP:journals/corr/abs-1801-06146}


Before large pre-trained networks like ViT, ResNet50 etc., neural networks were fine-tuned by training the latter layers of the network more as they are expected to be more specialized to the specific task the network was trained on, while the earlier layers are more task agnostic. Following this, the earlier layers are not adapted at all/only adapted slightly during transfer and the latter layers are adapted more freely. For the transformer architecture this practice is not widely adapted as in most papers Transformer based architectures are adapted equally across all layers.


In this paper we particularly introduce hyperparameter optimization technologies in this setting, which enables us to investigate a variety of learning rate distribution choices of the language model, during fine-tuning. To the best of our knowledge, this constitutes one of the first approaches in which the effect of specific learning rate on different parts of the transformer network on the fine-tuning process is extensively investigated.

\section{Proposed Approach}
\label{sec:req}

The transformer architecture is very powerful and is able to adapt to most classification tasks well, but is still susceptible to the problem of catastrophic forgetting \cite{ramasesh2022effect}. 

Here, we propose an approach to automatically select different learning rates for different parts of a transformer model in order to reduce the effect of catastrophic forgetting. Our approach is based on a two-step method: First, we determine such a learning rate distribution for a pair of two datasets with the goal to obtain the best performance for both while training sequentially. Then, we do this for a few such pairs and, in the second step, combine the resulting learning rate distributions such that catastrophic forgetting is also reduced for a novel unseen dataset pair. 

For the first step, we consider an original dataset $D_o$ and a shifted one $D_s$.
The indicated model is first fine-tuned on $D_o$ and then fine-tuned on $D_s$. Subsequently, the performance $p_o$ on $D_o$ is investigated for catastrophic forgetting, while the performance $p_s$ on $D_s$ is also computed to guide the hyperparameter optimization.
We combine the performance measures on both datasets to provide the rewards measure used during our hyperparameter optimization $p = p_s + p_o$.

The second step, the combination of different learning rate distributions is detailed in section \ref{chap:combining} and the search space for the hyperparameter optimization is described in section \ref{chap:searchspace}.

\subsection{Search Space}
\label{chap:searchspace}

In this paper we postulate that different learning rates 
on different parts of the network could provide value by letting these parts adopt at different speeds. The more general parts of the network will be modified less than the parts of the network which are responsible for more task specific computations. 
With this in mind we determine a space of 
 learning rate distribution choices, which constitute the possible search options.
We perform the choice of a different learning rate in the range of $1e-7$ to $1e-3$ for 10 different parts of the network, i.e.\ $lrs_j = \{ lr_{ij}, i=1, \dots, 10 \}, ~lr_{ij}\in[1e-7, 1e-3]$, for candidates enumerated with $j$. 

This enables the hyperparameter optimization to find a good configuration for every part of the network. We do not expect that 10 vastly different learning rates are needed, but rather choose this number high, as the optimization process has the possibility to converge to similar choices for consecutive parts of the network, thereby reducing the variation in the learning rates chosen.

Since BERT has 12 encoder layers we divide the different layers equally into 8 of these (choice 1 affects layer 1, choice 2 affects layers 2,3, choice 3 affect layer 4, choice 4 affect layers 5,6 ...).
Choice 0 is reserved for the embedding weights of the networks. Choice 9 is reserved for the single dense layer which is appended for a specific task.


The optimization strategy used is Bayesian Optimization \cite{bayesopt} with Hyperband \newline Scheduling \cite{li2017hyperband}, as a particularly promising technology in the domain of hyperparameter optimization.
We refer to the output of this procedure, i.e., the best configuration found by this approach in a specific training task, as BERT continual Learner or short $BERTcL$. 

\subsection{Combining Learning Rate Distributions} \label{chap:combining}

During the training on one dataset pair $a$, the optimization process evaluates different learning rate vectors $lrs^a_j = (lr^a_{j,1},\ldots,lr^a_{j,10})$ for layers $1,\ldots,10$ and experimental runs $j=1,\dots$. Each set of $lrs^a_j$ is evaluated and hereby assigned a performance measure $p^a_j$ resulting in a rank $r^a_j$ if sorted according to experimental performance $p^a_j$ for all $j$. 
Instead of using just the best performing learning rate set, we utilize a weighted combination of them for the purpose of generalization to new data. They are weighted by their performance ranking and combined with the weighted geometric mean (in correspondence to the exponential search of the learning rates):
\begin{equation}
    lr^{a}_i = \exp\left( \frac{1}{\sum_j b^{-r^a_j}}\sum_{j} \ln(lr_{ji}^a) * b^{-r_j^a} \right),  \mathrm{\ for \ all \ layers \ } i
    \label{eq:logavgweighted}
\end{equation}
with $b \in [1,\infty]$. We choose $b = 1.8$ (higher values put more emphasis on the best performing samples during averaging).

In practice performing hyperparameter optimization for every dataset pair is infeasible and not desirable. Consequently, we try to find a learning rate distribution which improves the performance for most dataset combinations. To do this, we combine the learning rate distributions $lrs^a$ found for a few dataset pairs $a$ using the geometric mean. 


We call the thusly created BERT version $BERTcL~ combined$.

\section{Experimental Approach}
\label{sec:experiments}

In this section, we detail our experimental design to investigate the effects of different learning rates for different parts of the BERT network for natural language classification tasks.
We utilize AutoGluon \cite{abohb} and the BertHugginface library \cite{wolf2020huggingfaces} for implementation and the pre-trained Bert model ('bert-base-uncased') for all experiments.  



We evaluate the two steps of our proposed approach: First we investigate in how far optimizing learning rates for different parts of the model can reduce catastrophic forgetting. For this purpose, we compare the performance drop when fine-tuning standard Bert subsequently to $BERTcL$ and EWC. This comprises a sanity check in how far our proposed modelling is in principle capable of reducing catastrophic forgetting. 
Then we perform the more interesting evaluation of the second step, whether $BERTcL~ combined$, i.e.\ a fixed optimized learning rate distribution for a specific pretrained model, is able to reduce catastrophic forgetting  for unseen datasets.

In each step, the training and testing is done on two independent subsets of the dataset. The splits as provided by the data source are used. We follow the EWC \cite{DBLP:journals/corr/KirkpatrickPRVD16} paper insofar that we change tasks or data subsets sequentially and evaluate the success of our method by the performance on the combination of performances at the end of training on all datasets.
Further details on the data and on hyperparameters are given in the following.

\subsection{Datasets}

The Glue dataset by Wang et al. \cite{wang2019glue} is a collection of various popular datasets in NLP, and it is widely used to evaluate common natural language processing capabilites. All datasets used are the version provided by tensorflow-datasets 4.0.1. 
More specifically, we use the Stanford Sentiment Treebank \emph{SST2}, the 
 Microsoft Research Paraphrase Corpus \emph{MRPC}, the Recognizing Textual Entailment \emph{rte}, the Stanford Question Answering Dataset \emph{QNLI}, the Quora Question Pairs2 Dataset \emph{QQP},
 and the Multi-Genre Natural Language Inference Corpus 
  \emph{MNLI}.

\subsection{Types of data shift}
In this paper we look at 2 different kinds of shift to evaluate the problem of catastrophic forgetting in transformer networks.

\noindent \textbf{Dataset shift}
With the term dataset shift we refer to a shift where the dataset and thereby task of the network shifts to a different task. By choosing the first and second task carefully we can produce more or less extreme shifts.
For a shift with a small impact, we choose the \emph{MRPC} dataset and the \emph{QQP} dataset, here the task of evaluating if two questions are the same shifts to the task of evaluating if two sentences are the same. 

For a more substantial shift, we evaluate the \emph{SST2} dataset and the \emph{MRPC} dataset, here the task changes from evaluating sentiment of a single sentence, to predicting entailment of 2 different sentences. This kind of shift shows if the transformer forgets part of the knowledge learned during pre-training and thus is no longer able to generalize to other tasks. \vspace{0.1cm}

\noindent \textbf{Distribution shift} 
Here, we talk about shifts where the task the network is trained upon stays the same, but rather some characteristics of the data distribution change.

\emph{Sentence length shift}
With the term sentence length shift we refer to a shift created by splitting the dataset by sentence length into 2 equally sized smaller datasets. One with sentences smaller than average $D_o$, the other with sentences larger $D_s$. Since the Transformer architecture is agnostic towards sequence length, this should have a comparatively small impact for the language model performance, but is still interesting.  

\begin{figure}[t]
    \centering
    \begin{subfigure}{.48\textwidth}
  \centering
  \includegraphics[width=.8\linewidth ]{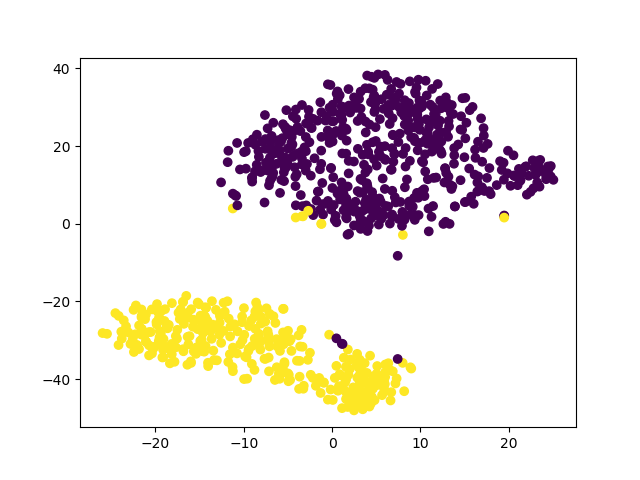}
  \caption{Embeddings of the \emph{SST2} dataset (validation data) colored according to their clusters computed with K-Means (K=2).}
  \label{fig:sfig1}
\end{subfigure}%
\hfill
\begin{subfigure}{.48\textwidth}
  \centering
  \includegraphics[width=.8\linewidth ]{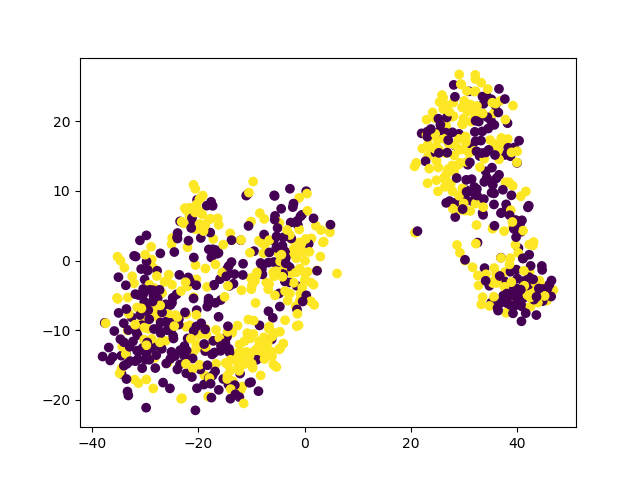}
  \caption{Embeddings of the \emph{SST2} dataset (validation data) colored according to their ground truth labels.}
  \label{fig:sfig2}
\end{subfigure}
\caption{Visualizations of sentences of the \emph{SST2} dataset embedded with BERT and projected to 2-D with TSNE.}
\end{figure}

\emph{Artificial shift}\label{sec:artshift}
We call the shift introduced by clustering the embeddings of the dataset generated by the pre-trained BERT into two clusters $D_o$ and $D_s$ with a clustering algorithm (K-means), artificial shift.
In the \emph{SST2} dataset the embeddings are naturally split in 2 clusters, see Figure \ref{fig:sfig1}, \ref{fig:sfig2}.

\subsection{Baselines and Implementation Details}

As a Baseline comparison we evaluate BERT with a learning rate of 2e-5, henceforth referenced to as $BERTbase$. These values are taken from the original paper \cite{bert} and present good values for a variety of classification tasks.

For another comparison we evaluate BERT combined with Elastic weight decay as a state of the art method to counter catastrophic forgetting \cite{DBLP:journals/corr/KirkpatrickPRVD16}, referenced as $BERT$ + $EWC$. The task used to compute the Elastic weight decay loss is the Masked Language Modeling task used during pre-training of BERT, computed on the Wiki40b \cite{49029} dataset. The importance value is set to 675 as indicated in \cite{DBLP:journals/corr/abs-2109-10021}.

The metaparameter choice for all experiments are as follows:
\begin{itemize}
    \item All models are trained for 5 epochs on the smaller (\emph{SST2}, \emph{RTE} and \emph{MRPC}) datasets and for 3 epochs on the larger (\emph{MNLI}, \emph{QNLI} and \emph{QQP}) datasets.
    \item All models are trained using a cosine decay of their learning rate with warm starting for 10 \% of the total training time.
    \item The pooling operation used in all experiments is [CLS]. 
    \item Batch size used for training was 16.
    \item The Adam optimizer with betas (0.9,0.999) and epsilon 1e-08 was used.
    \item The maximum sequence length was set to 256 tokens.
    \item No callbacks for early stopping were used. 
    \item A random seed (999) was used for numpy and python, not pytorch.
\end{itemize}
 

\section{Results}
In Table \ref{Fig:glue} we display the baseline results on the GLUE tasks without any datashift. Here we use the results obtained by the Huggingface BERT implementation and not the results of the original paper, as the original paper did not provide performances for all GLUE tasks.
\begin{table}[t]
  \centering
  \caption{Classification accuracies, without any shift for later comparison.}
  \label{Fig:glue}
  \begin{tabular}{c cccccc c}
    \toprule
method & SST2 &  MRPC & MNLI & QNLI & QQP & RTE & average \\
 \cmidrule(r){1-1} \cmidrule(l){2-7} \cmidrule(l){8-8}
$BERT base$ &  0.925 & 0.860 & 0.829 & 0.905 & 0.899 & 0.700 &  0.848 \\
    \bottomrule
  \end{tabular}
\end{table}

\begin{table}[b] 
  \centering
  \caption{Classification accuracies, for the dataset shift experiment. 
  The first number denotes the performance $p_o$ on the original dataset after training on the second dataset. The second number denotes the performance $p_s$ on the second dataset after training. Improvements are marked in \textbf{bold}. }
  \label{Fig:datashift}
  \begin{tabular}{c cccc c}
    \toprule
method & SST2-MRPC &  MRPC-SST2 & QQP-MRPC & MRPC-QQP &  average \\
 \cmidrule(r){1-1} \cmidrule(l){2-5} \cmidrule(l){6-6}
$BERT base$ &  0.904 - 0.860 & 0.740 - 0.922 & 0.869 - 0.831 & 0.725 - 0.901  &  0.809 - 0.879  \\
$BERT $ + $EWC$ & 0.845 - 0.836 & 0.650 - 0.915 & 0.851 - 0.855  & 0.691 - 0.895 &  0.759 - 0.875  \\

$BERTcL$ & \textbf{0.923} - 0.853 &  \textbf{0.765} - 0.910 & \textbf{0.877} - 0.838 & \textbf{0.767} - 0.892 &  \textbf{0.833} - 0.873 \\

$BERTcL$ combined & 0.908 - 0.836 &  0.755 - 0.916 & 0.877 - 0.823 & 0.738 - 0.891 & 0.820 - 0.867  \\
    \bottomrule
  \end{tabular}
\end{table}

\subsection{Dataset shift}

In the dataset shift experiments the performance drop off is quite significant depending on the datasets used, as can be seen in Table \ref{Fig:datashift}, compared to Table \ref{Fig:glue}. Our method of searching for a good learning rate distribution for every dataset combination called $BERTcL$ can mitigate this performance drop or in some cases, completely negate it. It results in on average 2.4\% better performance $p_o$ after training on the second dataset $D_s$, while having a small performance drop of $p_s$ of 0.6\% on the second dataset. 

\begin{figure}[t]
    \centering
    \includegraphics[width=.525\linewidth ]{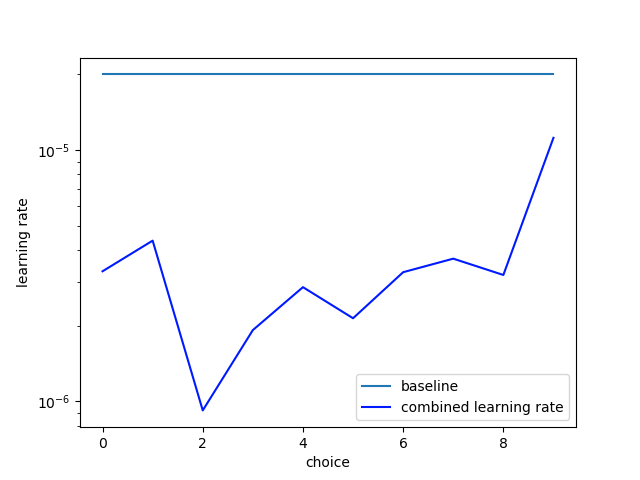}
    \caption{Combined learning rate as determined by the hyperparameter optimization process over the dataset shift experiments. X-Axis denotes position of learning rate in the transformer architecture as described in section \ref{chap:searchspace}. Lower numbers indicate earlier layers in the transformer. Y-Axis denotes the learning rate (log scale). }
    \label{fig:avglr}
\end{figure}

We combine the learning rate distributions found as described in Chapter \ref{sec:req} to create $BERTcL~ combined$.
The resulting learning rate distribution is visualized in Figure \ref{fig:avglr}. It is on average lower than the standard learning rate of $2e-5$, but shows a spike for the last (newly initialized dense) layer, as well as a very low learning rate for choice 2 (in the BERT architecture this represents the encoder layers 2 and 3). This suggest that these layers are very general and do not need to be retrained much for a specific task. 


For the sake of completeness, we provide $BERTcL~ combined$ scores on the datasets it was trained on in Table \ref{Fig:datashift}. The more interesting case of generalization towards unseen data during the hyperparameter optimization process is illustrated in Table \ref{Fig:datashiftunseen}. In all cases a better performance $p_o$ on the first dataset is achieved compared to $BERTbase$ with the combined distributed learning rate. However, a lower performance $p_s$ on the second dataset is also sometimes observed. Overall, on average over the unseen data, $BERTcL~ combined$ outperforms the flat learning rate, as well as $BERT$ + $EWC$ by about 5\%.

\begin{table}[b]
  \centering
  \caption{Classification accuracies for the dataset shift experiment on unseen datasets during the hyperparameter optimization process search. 
  The first number denotes the performance on the original dataset after training on the second dataset. The second number denotes the performance on the second dataset after training. Improvements are marked in \textbf{bold}.}
  \label{Fig:datashiftunseen}
  \begin{tabular}{c cccc c}
    \toprule
method & RTE-MRPC &  MRPC-RTE & MNLI-QNLI & QNLI-MNLI &  average \\
 \cmidrule(r){1-1} \cmidrule(l){2-5} \cmidrule(l){6-6}
$BERT base$ &  0.523 - 0.836 & 0.745 - 0.632 & 0.693 - 0.896  & 0.517 - 0.833  & 0.620 - 0.799  \\
$BERT $ + $EWC$ & 0.570 - 0.846 & \textbf{0.774} - 0.639 & 0.610 - 0.908 & 0.509 - 0.835 &   0.616 - 0.807 \\
$BERTcL$ combined & \textbf{0.588} - 0.760 &  0.760 - 0.599 & \textbf{0.770} - 0.893 & \textbf{0.595} - 0.815 & \textbf{0.678} - 0.767  \\
    \bottomrule
  \end{tabular}
\end{table} 

In conclusion, we find that $BERTcL~ combined$ constitutes a robust, flexible, easy to implement and well performing method to mitigate the problem of catastrophic forgetting during dataset shifts.

\subsection{Distribution shift}

In this section we depict results of the distribution shift experiments described in \ref{sec:artshift}.

\noindent \textbf{Sentence length shift}
For the sentence length shift experiments we choose the datasets \emph{SST2} and \emph{QNLI}. The results can be seen in Table \ref{Fig:distshift}. They indicate that the original BERT architecture can generalize well from longer to smaller sentences. Our $BERTcL$ is not expected to provide any improvement and we do not perform further evaluation on this kind of data shift. \vspace{0.1cm}


\begin{table}[t]
  \centering
  \caption{Classification accuracies, for the distribution shift experiment. Two numbers per dataset are given, the first denoting the accuracy on the partial dataset before the data shift and the second denoting the same accuracy after the data shift.}
  \label{Fig:distshift}
  \begin{tabular}{c c c c c c}
    \toprule
 & \multicolumn{2}{c}{sentence length shift} & \multicolumn{3}{c}{artifical shift} \\
 \cmidrule(r){1-1} \cmidrule(l){2-3} \cmidrule(l){4-6} 
method & SST2 &  QNLI & SST2 &  MRPC & MNLI  \\
 \cmidrule(r){1-1} \cmidrule(l){2-3} \cmidrule(l){4-6} 
$BERT base$ &  0.931 - 0.933 &  0.903 - 0.905 & 0.909 - 0.900 & 0.845 - 0.888  & 0.814 - 0.826 \\
    \bottomrule
  \end{tabular}
\end{table}

\noindent \textbf{Artifical shift}
The artifical shift experiment was performed as described in Section \ref{sec:artshift}, on the datasets \emph{SST2}, \emph{MRPC} and \emph{MNLI}. The results of these experiments can be seen in Table \ref{Fig:distshift}. The \emph{SST2} dataset has clearly seperable clusters, but the performance only drops slightly when introducing the artifical shift. For the \emph{MRPC} and \emph{MNLI} datasets the performance $p_o$ improves after training on $D_s$. Following, this we do not expect $BERTcL$ to provide any improvement.


In conclusion, we find that the transformer architecture is robust during distribution shifts, and can in many cases even improve performance, contrary to the catastrophic forgetting paradigm.



\section{Conclusion}
\label{sec:conclusion}

Transformer networks are surprisingly robust to varying lengths of input sequences during training and testing, as well as to artificially clustered data shifts.

For subsequent learning over consecutive datasets and tasks we present an intelligent learning rate distribution $BERTcL~ combined$ for the BERT sentence embedder which mitigates or in some cases completely solves the problem of catastrophic forgetting. 
In comparison to other approaches we do not modify the underlying transformer architecture or provide additional regularization, but rather change the learning rate based on the layer. This approach can be applied to many common encoder or decoder models like BERT, RoBERTa, distilBERT or GPT-2.

The exact contribution of the learning rate distribution found in comparison to a flat learning rate is of interest and merits further research. It is difficult to provide meaningful intuition about what changed in a transformer based network during training. While Attention visualizations can show on a case by case basis which words are associated with each other, this can only be done for individual sentences and is not representative of the dataset as a whole. 
Possible further research also includes other options for the hyperparameter optimization process, like different optimizer choices as well as learning rate schedules, for different parts of the network. 

\noindent
The source code is open-source and free (MIT licensed) software and available at \newline \url{https://github.com/TheMody/NAS-CatastrophicForgetting}.

\section*{Acknowledgements}
We gratefully acknowledge funding by the BMWi (01MK20007E) in the project AI-marketplace.

\bibliographystyle{splncs04}
\bibliography{references}

\begin{thebibliography}{10}
\providecommand{\url}[1]{\texttt{#1}}
\providecommand{\urlprefix}{URL }
\providecommand{\doi}[1]{https://doi.org/#1}

\bibitem{bert}
Devlin, J., Chang, M., Lee, K., Toutanova, K.: {BERT:} pre-training of deep
  bidirectional transformers for language understanding. CoRR
  \textbf{abs/1810.04805} (2018), \url{http://arxiv.org/abs/1810.04805}

\bibitem{49029}
Guo, M., Dai, Z., Vrandecic, D., Al-Rfou, R.: Wiki-40b: Multilingual language
  model dataset. In: LREC 2020 (2020),
  \url{http://www.lrec-conf.org/proceedings/lrec2020/pdf/2020.lrec-1.296.pdf}

\bibitem{HE2021106622}
He, X., Zhao, K., Chu, X.: Automl: A survey of the state-of-the-art.
  Knowledge-Based Systems  \textbf{212},  106622 (2021).
  \doi{https://doi.org/10.1016/j.knosys.2020.106622},
  \url{https://www.sciencedirect.com/science/article/pii/S0950705120307516}

\bibitem{DBLP:journals/corr/abs-1801-06146}
Howard, J., Ruder, S.: Fine-tuned language models for text classification. CoRR
   \textbf{abs/1801.06146} (2018), \url{http://arxiv.org/abs/1801.06146}

\bibitem{DBLP:journals/corr/abs-1911-03437}
Jiang, H., He, P., Chen, W., Liu, X., Gao, J., Zhao, T.: {SMART:} robust and
  efficient fine-tuning for pre-trained natural language models through
  principled regularized optimization. CoRR  \textbf{abs/1911.03437} (2019),
  \url{http://arxiv.org/abs/1911.03437}

\bibitem{DBLP:journals/corr/KirkpatrickPRVD16}
Kirkpatrick, J., Pascanu, R., Rabinowitz, N.C., Veness, J., Desjardins, G.,
  Rusu, A.A., Milan, K., Quan, J., Ramalho, T., Grabska{-}Barwinska, A.,
  Hassabis, D., Clopath, C., Kumaran, D., Hadsell, R.: Overcoming catastrophic
  forgetting in neural networks. CoRR  \textbf{abs/1612.00796} (2016),
  \url{http://arxiv.org/abs/1612.00796}

\bibitem{abohb}
Klein, A., Tiao, L., Lienart, T., Archambeau, C., Seeger, M.: Model-based
  asynchronous hyperparameter and neural architecture search. arXiv preprint
  arXiv:2003.10865  (2020)

\bibitem{DBLP:journals/corr/abs-2109-10021}
Kutalev, A., Lapina, A.: Stabilizing elastic weight consolidation method in
  practical {ML} tasks and using weight importances for neural network pruning.
  CoRR  \textbf{abs/2109.10021} (2021), \url{https://arxiv.org/abs/2109.10021}

\bibitem{DBLP:journals/corr/abs-1909-08383}
Lange, M.D., Aljundi, R., Masana, M., Parisot, S., Jia, X., Leonardis, A.,
  Slabaugh, G.G., Tuytelaars, T.: Continual learning: {A} comparative study on
  how to defy forgetting in classification tasks. CoRR  \textbf{abs/1909.08383}
  (2019), \url{http://arxiv.org/abs/1909.08383}

\bibitem{DBLP:journals/corr/abs-0712-3329}
Legg, S., Hutter, M.: Universal intelligence: {A} definition of machine
  intelligence. CoRR  \textbf{abs/0712.3329} (2007),
  \url{http://arxiv.org/abs/0712.3329}

\bibitem{li2017hyperband}
Li, L., Jamieson, K., DeSalvo, G., Rostamizadeh, A., Talwalkar, A.: Hyperband:
  A novel bandit-based approach to hyperparameter optimization. The Journal of
  Machine Learning Research  \textbf{18}(1),  6765--6816 (2017)

\bibitem{ramasesh2022effect}
Ramasesh, V.V., Lewkowycz, A., Dyer, E.: Effect of scale on catastrophic
  forgetting in neural networks. In: International Conference on Learning
  Representations (2022), \url{https://openreview.net/forum?id=GhVS8_yPeEa}

\bibitem{bayesopt}
Snoek, J., Larochelle, H., Adams, R.P.: Practical bayesian optimization of
  machine learning algorithms (2012). \doi{10.48550/ARXIV.1206.2944},
  \url{https://arxiv.org/abs/1206.2944}

\bibitem{DBLP:journals/corr/VaswaniSPUJGKP17}
Vaswani, A., Shazeer, N., Parmar, N., Uszkoreit, J., Jones, L., Gomez, A.N.,
  Kaiser, L., Polosukhin, I.: Attention is all you need. CoRR
  \textbf{abs/1706.03762} (2017), \url{http://arxiv.org/abs/1706.03762}

\bibitem{wang2019glue}
Wang, A., Singh, A., Michael, J., Hill, F., Levy, O., Bowman, S.R.: {GLUE}: A
  multi-task benchmark and analysis platform for natural language understanding
  (2019), in the Proceedings of ICLR.

\bibitem{wolf2020huggingfaces}
Wolf, T., Debut, L., Sanh, V., Chaumond, J., Delangue, C., Moi, A., Cistac, P.,
  Rault, T., Louf, R., Funtowicz, M., Davison, J., Shleifer, S., von Platen,
  P., Ma, C., Jernite, Y., Plu, J., Xu, C., Scao, T.L., Gugger, S., Drame, M.,
  Lhoest, Q., Rush, A.M.: Huggingface's transformers: State-of-the-art natural
  language processing (2020)

\end{thebibliography}

\end{document}